\documentclass[letterpaper, 10 pt, conference]{ieeeconf}  

\IEEEoverridecommandlockouts                              

\usepackage{graphics} 
\usepackage{epsfig} 
\usepackage{subfigure}
\usepackage{amsmath} 
\usepackage{epstopdf}
\usepackage{lipsum,amsmath,multicol}
\usepackage{float}
\usepackage{algorithm}
\usepackage{algorithmicx}
\usepackage{algpseudocode}

\usepackage{color,soul}
\definecolor{ilana}{rgb}{1, 1, 0}
\sethlcolor{ilana}
\setstcolor{ilana}

\title{\LARGE \bf
Stochastic Optimal Control for Modeling Reaching Movements in the Presence of Obstacles: Theory and Simulation
}

\author{Arun Kumar Singh$^{1,2}$, \emph{Member, IEEE}, Sigal Berman$^{1}$, \emph{Member, IEEE}, and Ilana Nisky$^{2}$, \emph{Member, IEEE}
\thanks{The research was supported by the Helmsley Charitable Trust through the Agricultural, Biological and Cognitive Robotics Initiative of Ben-Gurion University of Negev, Israel, and by the Israeli Science Foundation (grant 823/15)}
\thanks{$^{1}$ Department of Industrial Engineering and Management, Ben-Gurion University of the Negev, Israel
      }%
\thanks{$^{2}$ Department of Biomedical Engineering, Ben-Gurion University of the Negev, Israel
        }%
}

\begin{document}

\maketitle
\thispagestyle{empty}
\pagestyle{empty}

\allowdisplaybreaks

\begin{abstract}
In many human-in-the-loop robotic applications such as robot-assisted surgery and remote teleoperation, predicting the intended motion of the human operator may be useful for successful implementation of shared control, guidance virtual fixtures, and predictive control. Developing computational models of human movements is a critical foundation for such motion prediction frameworks. With this motivation, we present a computational framework for  modeling reaching movements in the presence of obstacles.  We propose a stochastic optimal control framework that consists of probabilistic collision avoidance constraints and a cost function that trades-off between effort and end-state variance in the presence of a signal-dependent noise. First, we present a series of reformulations to convert the original non-linear and non-convex optimal control into a parametric quadratic programming problem. We show that the parameters can be tuned to model various collision avoidance strategies, thereby capturing the quintessential variability associated with human motion. Then, we present a simulation study that demonstrates the complex interaction between avoidance strategies, control cost, and the probability of collision avoidance. The proposed framework can benefit a variety of applications that require teleoperation in cluttered spaces, including robot-assisted surgery. In addition, it can also be viewed as a new  optimizer which produces smooth and probabilistically-safe trajectories under signal dependent noise.
\end{abstract}

\section{Introduction}

Motion prediction is important for finding the middle ground between pure teleoperation and autonomous control of robotic systems. It allows the robot to anticipate the future motions of the users and, consequently, their intention, and assist them in performing a given task. To improve the performance of motion prediction algorithms, it is beneficial to ground the prediction in experimentally-validated computational models of human movement \cite{wolpert2000}. Optimal control is used extensively in computational motor control, and provides a powerful framework for explaining a wide range of empirical phenomena associated with human motion \cite{Flash1985, Flash2013, Todorov2004}. In this view, it is hypothesized that human motion is driven by well-defined rewards or cost functions. The complimentary
Inverse Optimal Control (IOC) framework attempts to identify the structure and parameters of these cost functions from a set of observed trajectories \cite{todorov_inverse}. Thus, IOC allows for the transition from modeling of human motion to motion prediction in a particular task \cite{mainprice_ioc}. However, the accuracy of the model that is used for a particular problem is critical to the success of IOC-based approaches.   While many studies considered optimal control for modeling reaching trajectories in free space \cite{Flash1985, Todorov2004, Scott2004, Diedrichsen2010},  there has been much less effort towards modeling reaching in the presence of obstacles using optimal control \cite{wolpert_obstacle}. This in turn, hinders the development of efficient IOC based approaches for prediction. 

In the current paper, we propose a stochastic optimal control framework for modeling human reaching trajectories in the presence of obstacles. This framework is designed to be incorporated in motion prediction for a variety of applications of teleoperation in cluttered spaces. Our proposed framework is built on experimental studies that suggest that reaching movements amongst obstacles are optimized considering the likelihood of collision \cite{chapman1, chapman2, mon-williams}, and that obstacle avoidance is sensitive to human perception of free space \cite{chapman2}. In line with these findings, the proposed optimal control model incorporates probabilistic collision avoidance constraints to ensure that the likelihood of collision is below a specified threshold. We also consider signal-dependent noise in human movement control \cite{Harris98}, and the uncertainty in the perception of the size of the obstacle to model the error in estimation of free space.

\noindent \textbf{Contributions:} Our main result is a reformulation of the optimal control problem proposed in \cite{wolpert_obstacle} which was shown to be effective in modeling reaching movements in the presence of obstacles. The proposed reformulations approximate a difficult non-linear and non-convex optimal control problem by a parametric quadratic optimization problem. We use substitution of chance constraints with a family of surrogate constraints \cite{bharath_iros15}. Satisfaction of each member of the family of surrogate constraints can be mapped to a lower bound probability with which the original chance constraints would be satisfied. Further, we show that the parameters of the reformulated quadratic optimization problem can be tuned to generate a diverse class of trajectories. To make the optimal control computationally tractable, we adapt \cite{Flash1985} and approximate the hand dynamics as a stochastic triple integrator system. Thus, our formulation does not address all the features of human reaching. Instead, we focus on capturing how  parameters of our optimal control model that represent risk seeking behavior of human can explain the trade-off between movement velocity and obstacle clearance in the vicinity of an obstacle.

The rest of the paper is organized as follows. Section \ref{rel} reviews the previous studies which considered collision avoidance within the context of optimal control. Section \ref{foc} presents the  optimal control problem followed by a series of reformulations to convert it into a tractable parametric quadratic optimization problem.  Section \ref{sim} presents simulation results that demonstrate how the parameters of the reformulated problem result in a diverse set of trajectories and control costs. In section \ref{disc} we discuss the results of our simulations in light of the existing experimental findings on reaching movements among obstacles and present future directions.

\section{Related Work}\label{rel}

\noindent \textbf{Optimal Control or Optimization based Obstacle Avoidance in Robotics}
Optimal control or optimization are used extensively to plan collision-avoiding trajectories that also optimize a specified cost function \cite{chomp}, \cite{stomp}. In \cite{stomp}, optimal control is applied to stochastic systems with additive noise, and collision avoidance is ensured by introducing a penalty on  trajectories that come close to the obstacles. An expectation over the cost is taken which suggests that the optimization is risk neutral; that is, it does not model the probability of collision avoidance. Trajectory optimizers like \cite{traj_opt}, \cite{muller2014risk} incorporates a penalty on the probability of collision avoidance. 

Some studies like  \cite{chance_avoid1}, \cite{chance_avoid2} put hard constraints on probability of collision avoidance . However, \cite{chance_avoid1}, \cite{chance_avoid2} assumed an additive noise model. We aim at planning trajectories for a human hand which is assumed to be modeled as a stochastic system with signal dependent noise \cite{Harris98}. An optimal control based framework presented in  \cite{vandenberg}  presents collision avoidance under signal dependent noise, but for single integrator systems. In contrast, our formulation incorporates a higher order dynamics.

\noindent \textbf{Obstacle Avoidance in Computational Motor Control}
Optimal control or optimization has been an important tool for studying arm movements in computational motor control community. These works include both deterministic \cite{Flash1985}, \cite{soechting}, \cite{kang}, \cite{uno} as wells as stochastic models \cite{Todorov2004}, \cite{Harris98}, \cite{tops},\cite{todorov_feedback}. Works like \cite{soechting}, \cite{kang}, \cite{uno} consider the full arm motion in their analysis. However, the arm dynamics are highly non-linear and its integration with probabilistic collision avoidance constraints would result in a computationally intractable optimal control problem. Thus, in contrast to these  works, we focus solely on the hand trajectories.

Reaching trajectories in the presence of obstacles were studied in computational motor control for understanding movement coordination. Experimental studies \cite{chapman1}, \cite{chapman2}, \cite{mon-williams}, \cite{Tressilian} investigated the effects of obstacle position and size on obstacle avoidance. In particular, \cite{Tressilian} observed that the obstacle avoidance strategy exhibited by human subject during reach to grasp movements, consisted of two basic  but coupled components namely moving around the obstacle or slowing down near them. 

An optimal control model for a single-obstacle avoidance was proposed in \cite{wolpert_obstacle}. They solved the optimal control problem using simulated annealing. Simple obstacle configurations, predominantly with a single obstacle were considered. Our proposed approach differs from \cite{wolpert_obstacle} in terms of the technical approach followed to solve the optimal control problem. In particular, we exploit some efficient structures in the problem. Moreover, we consider complex obstacle configurations to highlight the interaction between parameters, control cost and probability of collision avoidance. Our proposed approach also differs from \cite{todorov_obstacle} wherein obstacle avoidance is included as a cost function and consequently  do not model the probability of collision avoidance. Although, \cite{todorov_flexible} analyzes collision avoidance behavior in the presence of obstacles, the presented optimal control formulation do not explicitly include collision avoidance constraints or costs. Rather, collision avoidance is used as a test case to study variability of reaching movements as explained by stochastic optimal control as compared to other models.

\section{Proposed Forward Optimal Control (FOC)}\label{foc}

\subsection{Dynamics and Task Description}
We consider the task of reaching movements in a 2D cluttered environment. We chose a simple linear model for the movement of the end point of the hand -  a triple integrator -- system. We denote the state of the hand at time instant $t$ by $\textbf{X}^t = (x^t, y^t, \dot{x}^t, \dot{y}^t, \ddot{x}^t, \ddot{y}^t)$, where the individual state variables are defined as the Gaussian distributions. The parameters of the distributions, i.e. their means and variances are obtained from the following discrete time dynamics with jerk $U =(u_x,u_y)=(\dddot{x},\dddot{y})$ as the control input.

\small
\begin{equation}
X^{t+1} = \textbf{A} X^{t}+\textbf{B}(U^{t}+\varepsilon_{U}^t),
\label{linear}
\end{equation}
\normalsize
\noindent where $\textbf{A}$ and $\textbf{B}$ represent state transition and control scaling matrices of dimensions conforming to that of the state, and

\small
\begin{equation}
\varepsilon_U^t= \sum_{i=1}^{2}\phi_i \textbf{M}_iU
\label{varepsidef}
\end{equation}

\begin{equation}
\textbf{M}_1= \begin{bmatrix}
c_x     & 0  \\
0       & 0 \\
\end{bmatrix}, \textbf{M}_2 = \begin{bmatrix}
0       & 0  \\
0       & c_y \\
\end{bmatrix}.
\end{equation}  
\normalsize

The term $\varepsilon^{t}_U$ in (\ref{varepsidef}) represents the time varying signal-dependent noise, and is formulated in terms of constant scaling matrices $\textbf{M}_i$ and $\phi_i$ which are a set of zero-mean unit-variance normal random variables. This form of (\ref{varepsidef}) ensures that indeed the standard deviation of the noise grows linearly with the magnitude of the control signal \cite{Todorov2004}, and the constants $c_x,c_y$ determine the magnitude of noise as a fraction of the  control input.

\subsection{Optimal Control}
\noindent The discrete time optimal control can be represented by the following set of equations. 
\small
\begin{eqnarray}\label{opt1}
 \min J_{opt} = J_{U^t}+J_{X^t} \\\nonumber
Pr(C_j^t(x^{t},y^{t},x_j,y_j, R_j)\leq 0 )\geq \eta  , j = {1,2..n},\nonumber
\end{eqnarray}
\normalsize

\vspace{-0.4cm}

\small
\begin{equation}
J_{U^t} = \Vert U \Vert ^{2}, J_{X^t} = \sum_{t=t_0}^{t=t_f}E[L(X^{t},U^{t})],
\label{cont_statecost}
\end{equation}
\normalsize

\vspace{-0.3cm}

\small
\begin{equation}
L(X^{t},U^{t}) =  \sum_{i=1}^{6}w_i(X_i^t-X^{t_f})^2,
\end{equation}

\vspace{-0.3cm}

\begin{equation}
R_j \approx N(\mu_{R_j}, \sigma_{R_j}^2).
\label{pernoise}
\end{equation}
\normalsize

\noindent The objective function in (\ref{opt1}) consists of a control effort term and a state-dependent term which penalizes the end point variance of the trajectory. The term $w_i$ determines the relative weighting between the components of the state-dependent cost term. The constraints $C_j(.)\leq 0$ in  (\ref{opt1}) represent the collision avoidance requirement in a deterministic setting. Thus, the set of inequalities in (\ref{opt1}) signify  constraints  that the collision avoidance requirement is satisfied with a particular lower bound probability $\eta$.  The terms $x_j$, $y_j$ and $R_j$ denote the position and size of the $j^{th}$ obstacle. To model the uncertainty in the estimation of obstacle size, $R_j$ is defined as normally-distributed random variable. 

The optimization (\ref{opt1}) is difficult to solve due to the constraints on probability of collision avoidance, also known as chance constraints, and are computationally intractable \cite{chance1}. Hence, we next reformulate these chance constraints into a tractable form and show that the reformulation naturally leads to an efficient optimization structure.

\noindent \textbf{Reformulating Chance Constraints:} We follow \cite{bharath_iros15}, and substitute of $Pr(C_i^t(.))$  with:

\vspace{-0.2cm}

\small
\begin{eqnarray}\label{meanvar}
Pr(C_j^t(x^{t},y^{t},x_j,y_j, R_j)\leq 0 )\geq \eta\\\nonumber
\Rightarrow E[C_j^t(.)]+k\sqrt(Var[C_j^t(.)]\leq 0, \eta \geq \frac{k^2}{1+k^2}.
\end{eqnarray}
\normalsize

\noindent where $E[C_j^t(.)]$ and $Var[C_j^t(.)]$ represent the expectation and variance of the constraints $C_j^t(.)$ with respect to the random variables $x^t,y^t$. This suggests that satisfaction of the deterministic surrogate in \ref{meanvar} ensures satisfaction of the original probabilistic constraints with at least a probability $\frac{k^2}{1+k^2}$. In \cite{bharath_iros15}, it is shown that computing an analytical expression for $E[C_j^t(.)]$ and $Var[C_j^t(.)]$ in terms of random variable arguments $x^t,y^t,R_j$ etc. is simpler compared to computing that for $Pr(C_j^t(.))$. We can further simplify (\ref{meanvar}) by approximating obstacle regions in 2D as circles. This simplifies the collision avoidance inequality $C_j^t(.)$:

\small
\begin{equation}
C_j^t: -(x^t-x_j)^2-(y^t-y_j)^2 +R_j^2\leq 0.
\label{C_i}
\end{equation}
\normalsize

\noindent Because (\ref{C_i}) is purely concave in terms of hand position variables $x^t$ and $y^t$, an affine upper bound can by obtained by linearizing $C_i^t$ around an initial trajectory guess $(x_*^t,y_*^t)$ \cite{sqp}:

\vspace{-0.3cm}
\small
\begin{equation}
C_j^t\approx ^*C_j^t+\bigtriangledown_{x^t} C_j^t(x^t-x_*^t)+\bigtriangledown_{y^t} C_j^t(y^t-y_*^t)\leq 0,
\label{affine}
\end{equation}
\normalsize

\noindent Where, $^*C_j^t$ is obtained by evaluating (\ref{C_i}) at $(x_*^t,y_*^t)$. Similarly, $\bigtriangledown_{x^t} C_j^t $ and $\bigtriangledown_{y^t} C_j^t$ represent the partial derivative of $C_j^t(.)$ with respect to $x^t$ and $y^t$,  evaluated at $(x_*^t,y_*^t)$. The affine approximation (\ref{affine}) can be further improved by updating $(x_*^t,y_*^t)$, during the course of the optimization. This sequential linearization of concave constraints forms the basis of the \emph{convex concave procedure} \cite{sqp}.

\noindent In light of (\ref{affine}), $E[C_j^t(.)]$ and $Var[C_j^t(.)]$ take the   form

\small
\begin{eqnarray}\label{expec}
E[C_j^t(.)] = \sigma^2_{R_j}\\\nonumber+h_1(\mu_{x^t},x^t_*,\mu_{y^t},,y^t_*,\sigma_{x^t}^2, \sigma_{y_t}^2\mu_{x_j},\mu_{y_j},\mu_{R_j})
\end{eqnarray}
\normalsize

\vspace{-0.4cm}

\small
\begin{eqnarray}\label{var}
Var[C_j^t(.)] = C_{R_j}\sigma_{R_j}^2+2\sigma_{R_j}^4\\\nonumber
+h_2(\mu_{x^t},x^t_*,\mu_{y^t},y^t_*,\sigma_{x^t}^2,\sigma_{y^t}^2,\mu_{x_j},\mu_{y_j},\mu_{R_j}),
\end{eqnarray}
\normalsize

\noindent where the terms $(\mu_{x^t},\mu_{y^t})$ and $(\sigma_{x^t}^2,\sigma_{y^t}^2)$ represent the mean and variance of the hand position $(x^t,y^t)$. The term $C_{R_j}$ and functions $h_1(.)$ and $h_2(.)$ are given in (\ref{cri})-(\ref{h2}). It can be noted that $h_2(.)$ can be represented as sum of squares and thus, is non-negative.

\small
\begin{eqnarray}
C_{R_i} = 4\mu_{R_i}^2\label{cri}\\
h_1 = \mu_{R_i}^2+2\mu_{x^t}\mu_{x_i}-\mu_{x_i}^2+2\mu_{y^t}\mu_{y_i}-\mu_{y_i}^2-2\mu_{x^t} x^t_*\\\nonumber-2\mu_{y^t}y^t_*+(x^t_*)^2+(y^t_*)^2\label{h1}\\
h_2 = 2(2\mu_{x_i}^2\sigma_{x^t}^2+2\mu_{y_i}^2\sigma_{y^t}^2-4\mu_{x_i}\sigma_{x^t} ^2(x^t_*)^2-4\mu_{y_i}\sigma_{y^t}^2(y^t_*)^2\label{h2}\\\nonumber + 2\sigma_{x^t}^2(x^t_*)^2+2\sigma_{y^t}^2(y^t_*)^2)
\end{eqnarray}
\normalsize

\noindent \textbf{Reformulated Optimal Control Problem:} To arrive at the final reformulated version of (\ref{opt1}), we make the following sequence of observations. The second term of the surrogate constraints proposed in (\ref{meanvar}) is non-negative. Thus, for a given $k$, the surrogate constraints (\ref{meanvar}) are satisfied when the first term, $E[C_j^t(.)]$ is sufficiently negative and the second term, $\sqrt(Var[C_j^t(.)]$ is sufficiently small in magnitude. Due to (\ref{var}) and (\ref{h2}) we note that  $\sqrt(Var[C_j^t(.)]$ is a non-decreasing function of the positional variance at each point of the trajectory $(\sigma^2_{x^t},\sigma^2_{y^t})$. Thus, making $\sqrt(Var[C_j^t(.)]$ small is equivalent to minimizing the positional variance at each point of the trajectory. In light of all these arguments, FOC (\ref{opt1}) can be replaced with the following simpler problem.

\vspace{-0.3cm}
\small
\begin{eqnarray}\label{opt2}
 J_{aug}= \Vert U \Vert ^{2}+\sum_{t=t_0}^{t=t_f}E[L(X^{t},U^{t})]+\lambda\sum_{t= t_0}^{t_f}(\sigma^2_{x^t}+\sigma^2_{y^t})\\\nonumber
E[C_j^t(.)]+\tau \leq 0
\end{eqnarray}
\normalsize

The original trajectory optimization (\ref{opt1}) has been converted to the new formulation (\ref{opt2}) by substituting the parameter $\eta$ which represented probability of avoidance in (\ref{opt1}) with two new sets of variables $\tau$ and $\lambda$. The positive constant $\tau$ can be manipulated to make $E[C_j^t(.)]$ as negative as required and consequently control the clearance from a given set of obstacles. Similarly, $\lambda$ is a positive constant which can be manipulated to minimize the positional variance at each point along the trajectory. Hence, we can manipulate $\tau$ and $\lambda$ to achieve a particular probability of avoidance $\eta$. Moreover, each  $\eta$ can be mapped to various choices of $\tau$ and $\lambda$ leading to a diverse set of collision avoidance behaviors. Within this diverse set, $\tau$ determines the geometry of the path, and $\lambda$ determines the velocity profile along the path. 

The reformulated FOC (\ref{opt2}) is very different from those typically used in the context of human motion modeling. A central hypothesis in current frameworks is that relative weighting of each term in the cost function can be tuned to produce a diverse set of trajectories. The FOC (\ref{opt2}) takes on a different approach -- its parameters appear not only in the cost function but also in the constraints.

The reformulated FOC (\ref{opt2}) can be solved in one shot if the right set of $\tau$ and $\lambda$ are given. For the cases where such set is not available, we can derive a framework for mapping a probability of collision avoidance $\eta$ to $\tau$ and $\lambda$ and solving (\ref{opt2}) in the process.

\noindent \textbf{Solutions in Different Homotopies:} The linearization of collision avoidance constraints (\ref{C_i}) to obtain affine inequalities (\ref{affine}) inherently limits the solution trajectories of (\ref{opt2}) to be locally optimal. The physical interpretation of this is that FOC, (\ref{opt2}) cannot search over the solution trajectories belonging to different homotopies. Existing optimal control approaches capable of searching over different homotopies either reformulate collision avoidance constraints, (\ref{C_i}) through use of binary variables \cite{milp1} or introduce additional constraints which model the topological information about the different possible homotopies \cite{homotopy1}, \cite{homotopy2}, \cite{homotopy3}. However, adopting such approaches would significantly increased the complexity of our optimization, (\ref{opt2}). Instead, we opt for an approximate solution. We vary the initial trajectory guess to produce optimal trajectories in different homotopies. In particular, an initial guess for each homotopy is pre-computed and stored and recalled as and when required. This initial guess could be computed from even sampling based planners. Some existing works on stochastic optimal control based collision avoidance also adopt similar approach \cite{vandenberg_coll}. Our approximate approach is also motivated by our eventual future goal of using the proposed formulation for learning reaching movements. In that context, a data set of initial guesses in different homotopies can be obtained from the user demonstration.

\noindent \textbf{Efficiently Solving the Proposed FOC:}

\noindent Algorithm 1 summarizes a sequential quadratic programming (SQP) routine for  solving FOC (\ref{opt2}). The optimization starts with an initial guess trajectory $(x^t_*,y^t_*)$ and initialization of an index counter $i$ and two non-negative variables $\tau$ and $\lambda$. The outermost loop checks whether the constraints are satisfied and  reduction in the cost function between two consecutive iterations is within a specified threshold, $\xi$. If either of these checks are violated, then the algorithm proceeds to the inner loop where we check whether the surrogate constraints (\ref{meanvar}) are satisfied. If not, then we increment the value of the $\tau$ by $\delta$ and $\lambda$ by a factor of $\Delta$. Thereafter (\ref{opt2}) is solved with the current values of $\tau$ and $\lambda$ and the solution obtained is used to update the initial guess trajectory, which in turn is used to obtain a better estimate of $C_j^i(.)$ through (\ref{affine}) for the next iteration.

Algorithm 1 has two important features. Firstly, $E[C_j^t(.)]$ is affine and $J_{aug}$ is convex quadratic in terms of control variables. Thus, solving (\ref{opt2}) for a given $\tau$ and $\lambda$ amounts to solving a quadratic programming (QP) problem. This is turn can be accomplished efficiently through open source solvers like CVX \cite{CVX}. Secondly, algorithm 1 is different from the standard SQP routines used to solve general non-convex problems in the sense that it does not require a trust region update. This, in turn, is because the affine approximation of $C_j^t$ in (\ref{affine}) acts as a global upper bound for the original collision  constraints (\ref{C_i})

Each $\eta$ can be mapped to numerous combinations of $\tau$ and $\lambda$. This redundancy is captured in algorithm \ref{algo1} by manipulating the update rates of $\tau$ and $\lambda$. We discuss this in more detail in Section \ref{sim} with the help of specific examples.

\begin{algorithm}
 \caption{Sequential Quadratic Programming for solving FOC }\label{algo1}
    \begin{algorithmic}
    \small
    \State  \textbf{Initialization}: Initial guess for optimal trajectory $x^t_*,y^t_*$. 
    \State $i  =0$ ,$\tau = 0$, $\lambda = 1$
    \State \While  $\vert J^{i+1}_{opt}-J^i_{opt}\vert<\xi$ and $E[C_j^t(.)]+k\sqrt(Var[C_j^t(.)]\leq 0$
    \If{$E[C_j^t(.)]+k\sqrt(Var[C_j^t(.)]>0$}\\
    $\tau \leftarrow \tau+\delta$\\
    $\Delta \leftarrow \Delta\lambda$\\
    \EndIf\\
   $U$ $\leftarrow$ $ \arg\min J_{aug}$ \\\hspace{0.4cm} $E[C_j^t(.)]+\tau\leq 0$
   \State Update $x^t_*,y^t_*$ through $U$\\
   $i \leftarrow i+1$

    \EndWhile
    \normalsize
        \end{algorithmic}
        \end{algorithm}

\section{Simulation Results}\label{sim}

\subsection{Collision Avoidance Strategies}
To ensure collision avoidance, humans can choose to maintain high clearance from the obstacles resulting in a large deviation from straight line paths. Alternatively, they  can choose to reduce the deviation but compensate for it by moving with high precision near the obstacles (reduce positional variance).  In light of the the signal dependent noise (\ref{varepsidef}),  moving with precision near the obstacle requires moving with low velocities. For the ease of exposition, from hereon, we will refer to the slowing down strategy as "Low Velocity" or \textbf{LV} and strategy of maintaining large clearance from the obstacles as "High Clearance" or \textbf{HC}.

\begin{figure}[!h]
	
  \centering
   \subfigure[]{
    \includegraphics[width= 4.3cm, height=3.2cm] {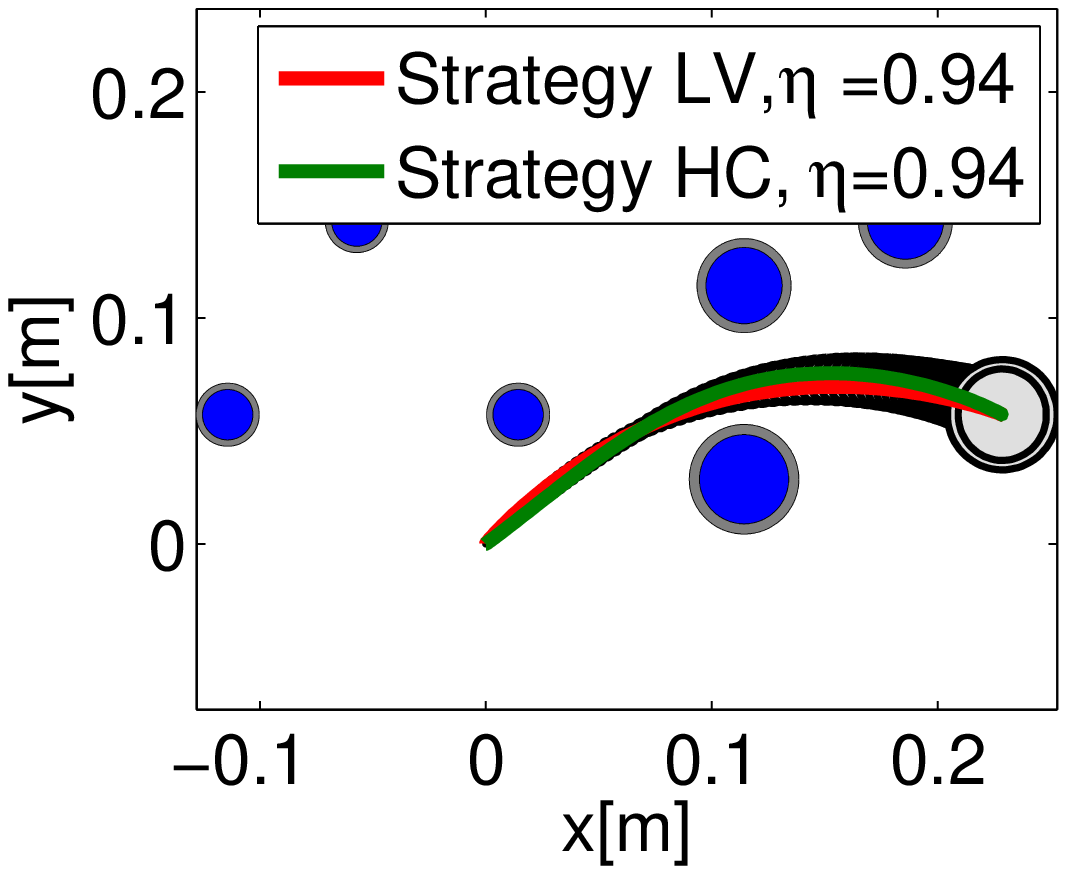}
    \label{plotcomp_dev_config1}
   }\hspace{-0.6cm}
   \subfigure[]{
    \includegraphics[width= 4.3cm, height=3.2cm] {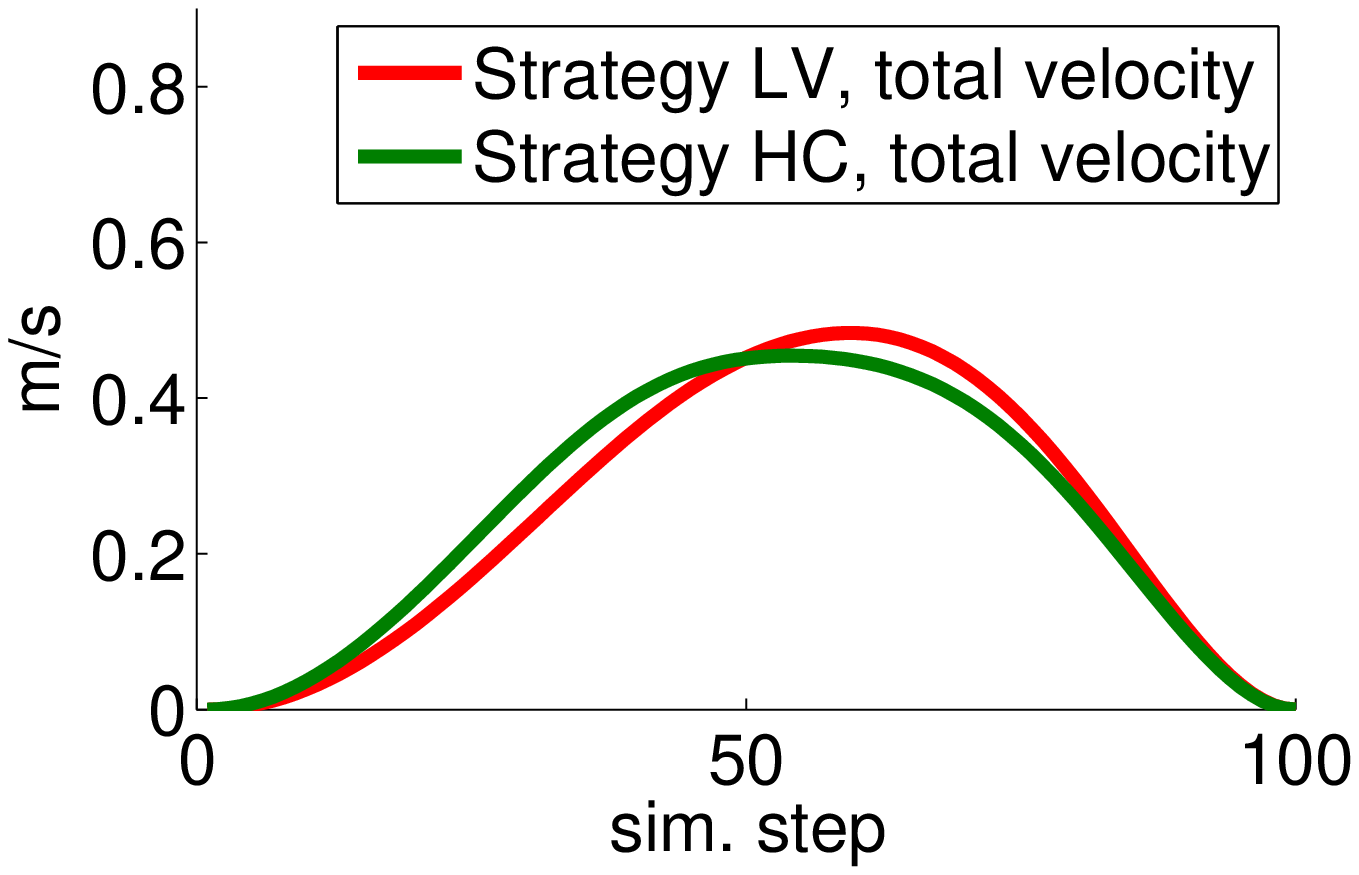}
    \label{plotcomp_dev_vel_config1}
   }\hspace{-0.8cm}
   \caption{Demonstration of the effect of the choice of $\tau$ and $\lambda$ on the collision avoidance strategies. Two sets of trajectories between same start and goal locations and having same probability of avoidance, $\eta$ were computed.  However, to generate these two trajectories we used a different set of $\tau$ and $\lambda$ to achieve the specified probability of avoidance. The trajectories shown in green were computed using $\tau=0.0012, \lambda = 0.9*10^6$, while trajectories shown in red were computed using $\tau=0.0009, \lambda = 2.28*10^6$.}     
    \label{plotcomp_config1}  
\end{figure}

Both these strategies can be modeled through (\ref{opt2}) by using appropriate values for parameters $\tau$ and $\lambda$. For example,   Fig.~\ref{plotcomp_config1} shows two solution trajectories of (\ref{opt2}) between the same start and goal configurations. The probability of avoidance, $\eta$ for both trajectories is $0.94$. However, both trajectories achieve this probability of collision avoidance through different combinations of $\tau$ and $\lambda$.  The trajectory resulting from strategy \textbf{LV} was obtained with  $\tau=0.0009, \lambda = 2.28*10^6$, while that resulting from strategy \textbf{HC} was obtained with $\tau=0.0012, \lambda = 0.9*10^6$. These values of $\tau$ and $\lambda$ were obtained using different update rates of of $\tau$ and $\lambda$ in algorithm \ref{algo1}. For simulating strategy \textbf{LV} we used $\delta = 0.00005$, $\Delta = 10$ in the update rule of $\tau$ and $\lambda$, and for simulating strategy \textbf{HC} we used $\delta = 0.0001$, $\Delta = 10$. Since, $\tau$ controls the clearance from the obstacles, setting higher update rates for $\tau$ resulted in trajectories belonging to strategy \textbf{HC}. On the other hand, a lower update rate for $\tau$  puts a higher emphasis on $\lambda$ and consequently manipulation of positional variance through velocity control for collision avoidance, thus, resulting in trajectories belonging to strategy \textbf{LV}.

The velocity profiles shown   in Fig.~\ref{plotcomp_dev_vel_config1} demonstrate that a higher $\lambda$ forces the velocity magnitude along the trajectory closer to the obstacle (strategy \textbf{LV}) to be small during the initial stages, i.e, while the trajectory is near the obstacles. Consequently, the positional variance is reduced and desired probability of collision avoidance is maintained. In contrast, the trajectory with higher clearance from the obstacle (strategy \textbf{HC}) has the liberty to move with faster velocity and let the variance of the movement grow. The velocity magnitude along trajectory resulting from strategy \textbf{LV} increases eventually, but this happens towards the end of the movement, after crossing the obstacles.

\subsection{Mapping Avoidance Strategies to Control Cost}

If we would derive a variant of the optimization (\ref{opt2}) for a system with an additive constant-variance noise, the probability of collision avoidance, $\eta$ would solely depend on the clearance from the obstacles. Thus, increase in $\eta$ would directly lead to an increase in arc lengths, and consequently, control costs. However, to develop a framework that is suitable for modeling human arm movements,  we incorporated signal dependent noise \cite{Harris98}. In the presence of signal-dependent noise, control cost of trajectories depends on the probability of avoidance $\eta$, and more importantly, on the combination of $\tau$ and $\lambda$ that is used in the optimization (\ref{opt2}) to achieve this $\eta$. In other words, the control cost depends on the strategy that is used to achieve a particular probability of collision avoidance. 

\begin{figure}[!h]
  \centering
   \subfigure[]{
    \includegraphics[width= 4.4cm, height=3.2cm] {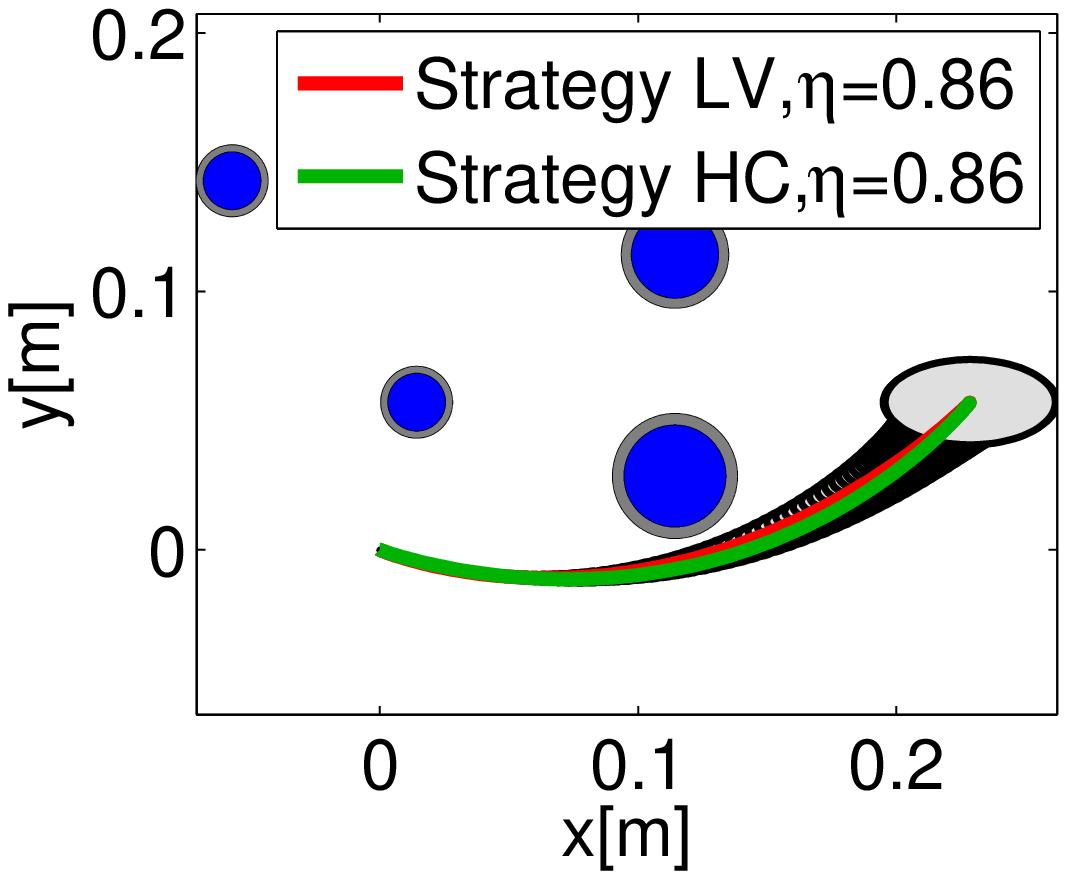}
    \label{strat_comp_traj1}
   }\hspace{-0.8cm}
   \subfigure[]{
    \includegraphics[width= 4.4cm, height=3.2cm] {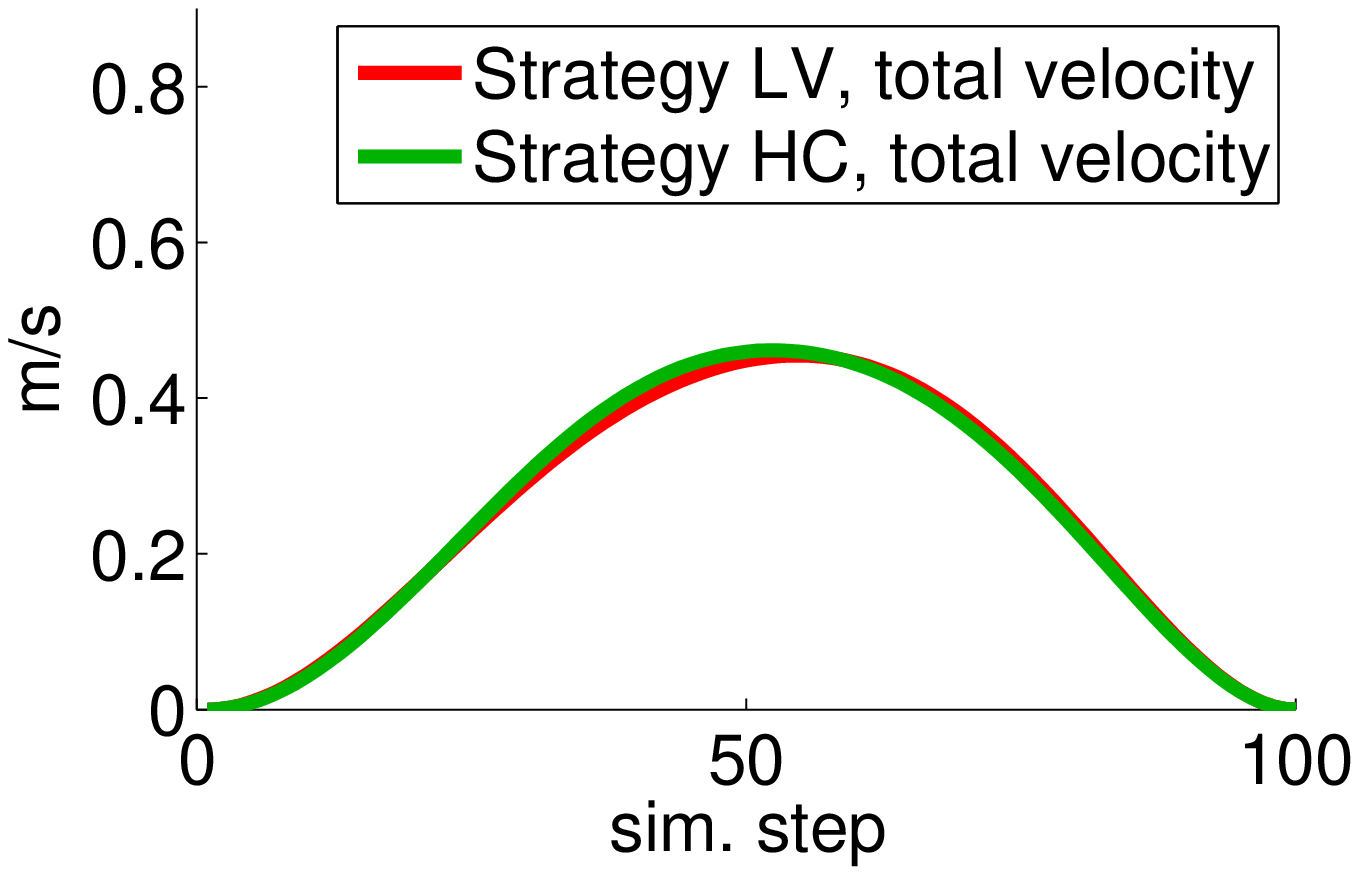}
    \label{strat_comp_traj1_vel}
   }\vspace{-0.7cm}
   \subfigure[]{
    \includegraphics[width= 4.4cm, height=3.2cm] {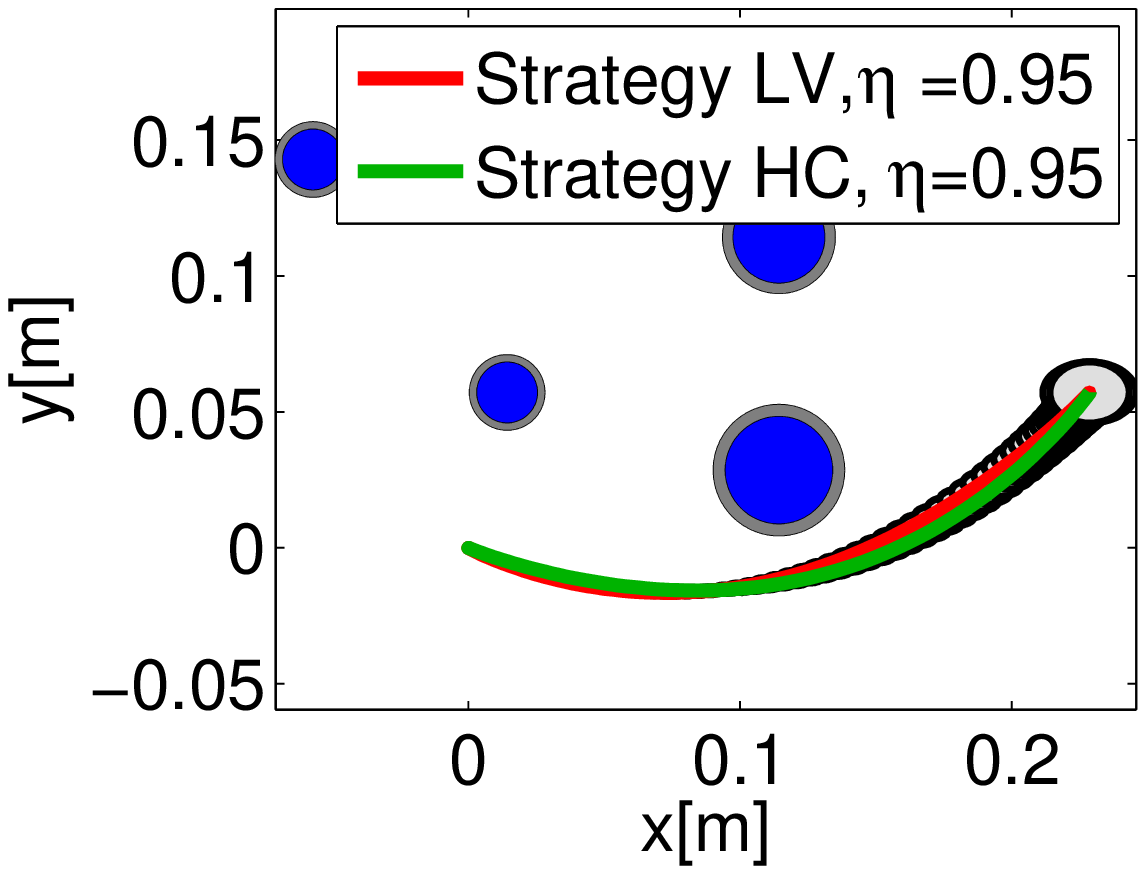}
    \label{strat_comp_traj2}
   }\hspace{-0.8cm}
   \subfigure[]{
    \includegraphics[width= 4.4cm, height=3.2cm] {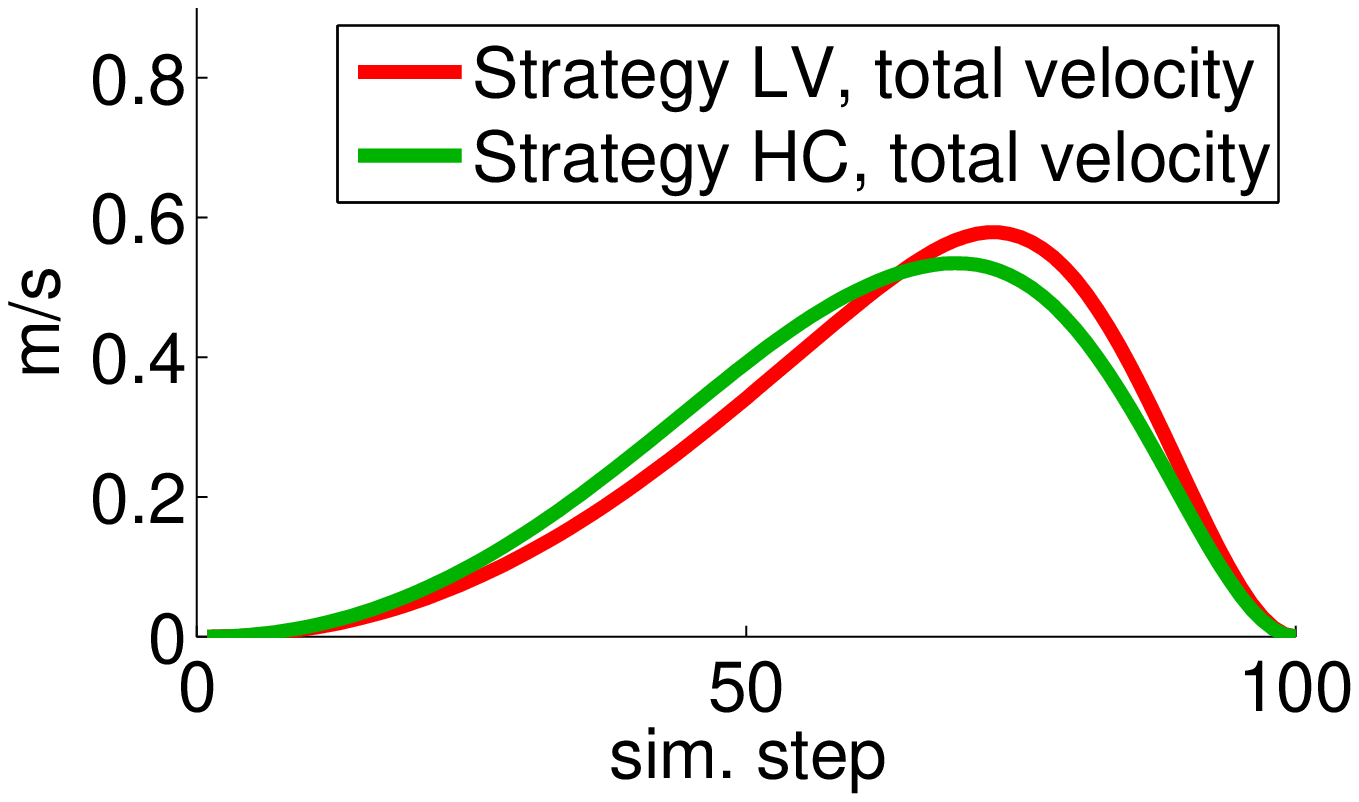}
    \label{strat_comp_traj2_vel}
   }\vspace{-0.7cm}
   \subfigure[]{
    \includegraphics[width= 8.3cm, height=4.3cm] {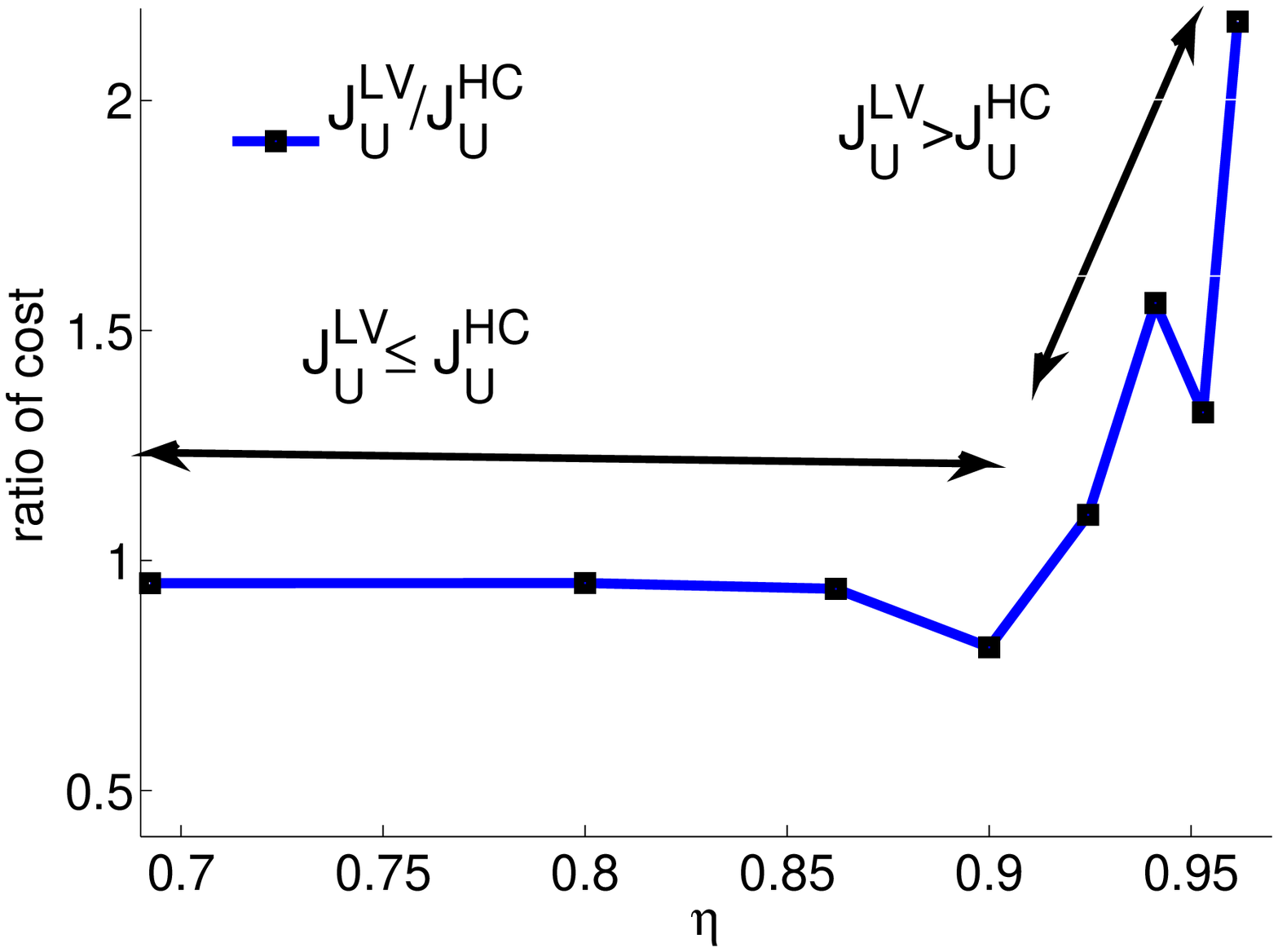}
    \label{cost_intra_hom}
   }
   \caption{Control costs vary with probability of avoidance. (a)- (d) Movements with different strategies between the same start and goal locations, the same obstacle configurations, and with noise level $c_x, c_y=0.15$. (a), (c) present the paths with standard deviation ellipses of the two strategies. The obstacles are represented as blue filled circles and grey shaded region around them represent uncertainty about the size of the obstacle. (b), (d) present the velocity profiles.  (e) the ratio of the control costs between the two strategies, $\frac{J_U^{LV}}{J_U^{HC}}$, as a function of $\eta$.}   
\vspace{-0.5cm}          
\end{figure}

In Fig. \ref{strat_comp_traj1}-\ref{strat_comp_traj2_vel} we present simulated trajectories that correspond to both strategy \textbf{LV} and \textbf{HC} for probabilities of collision avoidance $\eta=0.86$ and $\eta=0.95$. The paths that resulted from strategy \textbf{HC} indeed has higher clearance from the obstacles. In contrast, the paths that resulted from strategy \textbf{LV} have lower clearance and thus, heavily rely on modifying the velocity magnitudes and consequently positional variance for collision avoidance. Consequently, paths resulting from strategy \textbf{HC}  have higher arc lengths as compared to paths resulting from strategy \textbf{LV}.  In Fig.~ (\ref{cost_inter_hom}) the ratio of control costs for trajectories resulting from both the strategies is presented as a function of $\eta$. For low $\eta$, paths resulting from strategy \textbf{LV} which have lower arc lengths are less costly. But, as $\eta$ increases, the higher arc length paths resulting from strategy \textbf{HC} become less costly. 

The observations discussed above are apparent from the structure of the optimization (\ref{opt2}). Increasing either $\tau$, $\lambda$, or both, leads to an increase in the control cost. At low values of $\eta$, there is very little restriction on the growth of positional variance and thus the control cost is dictated by $\tau$ which controls the arc length. But as $\eta$ increases, the effect of $\lambda$ becomes prominent. This is consistent with the significant reduction in positional variance that is depicted in Fig.~ \ref{strat_comp_traj2} and the corresponding skewed velocity profile shown in Fig.~ \ref{strat_comp_traj2_vel}. Since trajectories resulting from strategy \textbf{LV} has lesser clearance from the obstacles, they require a higher value of $\lambda$ to achieve the same $\eta$ (similar to the result shown in previous section). Thus, at higher probabilities trajectories resulting from strategy \textbf{LV} become more costly in spite of having lower arc lengths.

\begin{figure}[!h]
\centering
\includegraphics[width= 8.3cm, height=3.7cm] {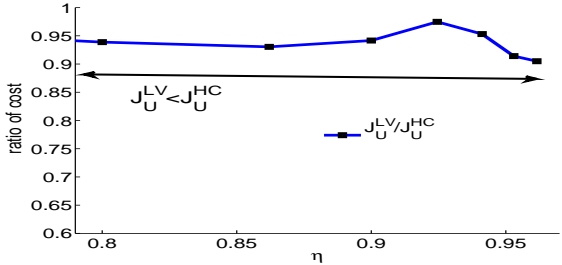}
\caption{ The ratio of the control costs between the two strategies, $\frac{J_U^{LV}}{J_U^{HC}}$, as a function of $\eta$ for noise level of $c_x, c_y=0.05$. }
\label{cost_lessnoise}
\vspace{-0.4cm}
\end{figure}

The results presented above, were obtained with $c_x=c_y=0.15$ in (\ref{varepsidef}). That is, the noise was $15\%$ of the control input. Next, we examined how the cost shown in Fig.~ \ref{cost_intra_hom} changes with a reduction in noise. Fig.~\ref{cost_lessnoise} depicts the ratio of control costs for trajectories resulting from strategy \textbf{LV} and \textbf{HC} for $c_x=c_y=0.05$. With lower noise, strategy \textbf{LV} becomes less costly even for higher probabilities. This result agrees with the common intuition. With a lesser noise there is no need to ensure high clearance from the obstacles, thereby making strategy \textbf{HC} redundant. In fact, for a zero noise system, the trajectory with least control cost would just graze the obstacle.

We would like to highlight that Fig.~ \ref{cost_intra_hom} and Fig.~\ref{cost_lessnoise} are intended to demonstrate the general trend in ratio of control costs. An in depth analysis of the exact values and their dependence on the initial conditions of the optimization are beyond the scope of this current study.

\subsection{Modeling Choice of Homotopies}

In this section, we  discuss how choice of strategy of collision avoidance or in other words, choice of $\tau$ and $\lambda$ for a given $\eta$ affects control cost of trajectories in different homotopies. 

\subsubsection{Strategy \textbf{LV}}
\begin{figure}[!h]
  \centering
   \subfigure[]{
    \includegraphics[width= 4.6cm, height=3.5cm] {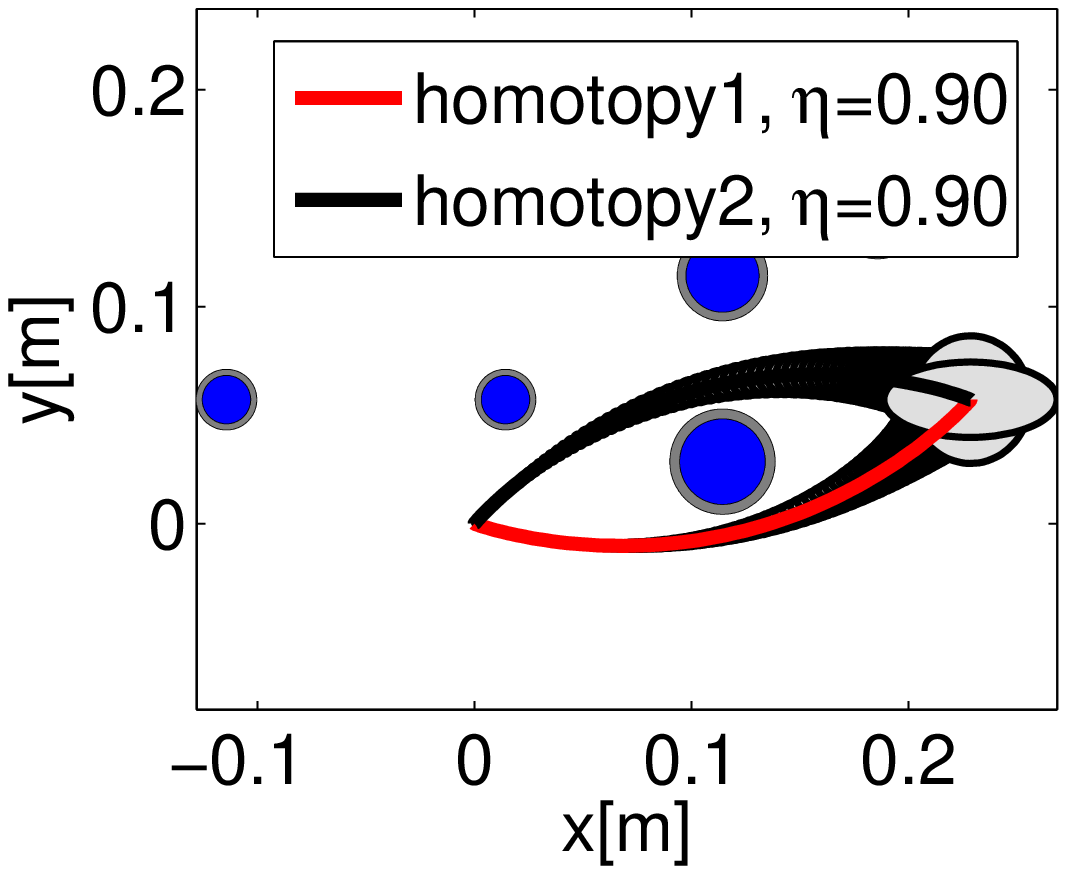}
    \label{plot1comp_config1}
   }\hspace{-0.8cm}
   \subfigure[]{
    \includegraphics[width= 4.1cm, height=3.5cm] {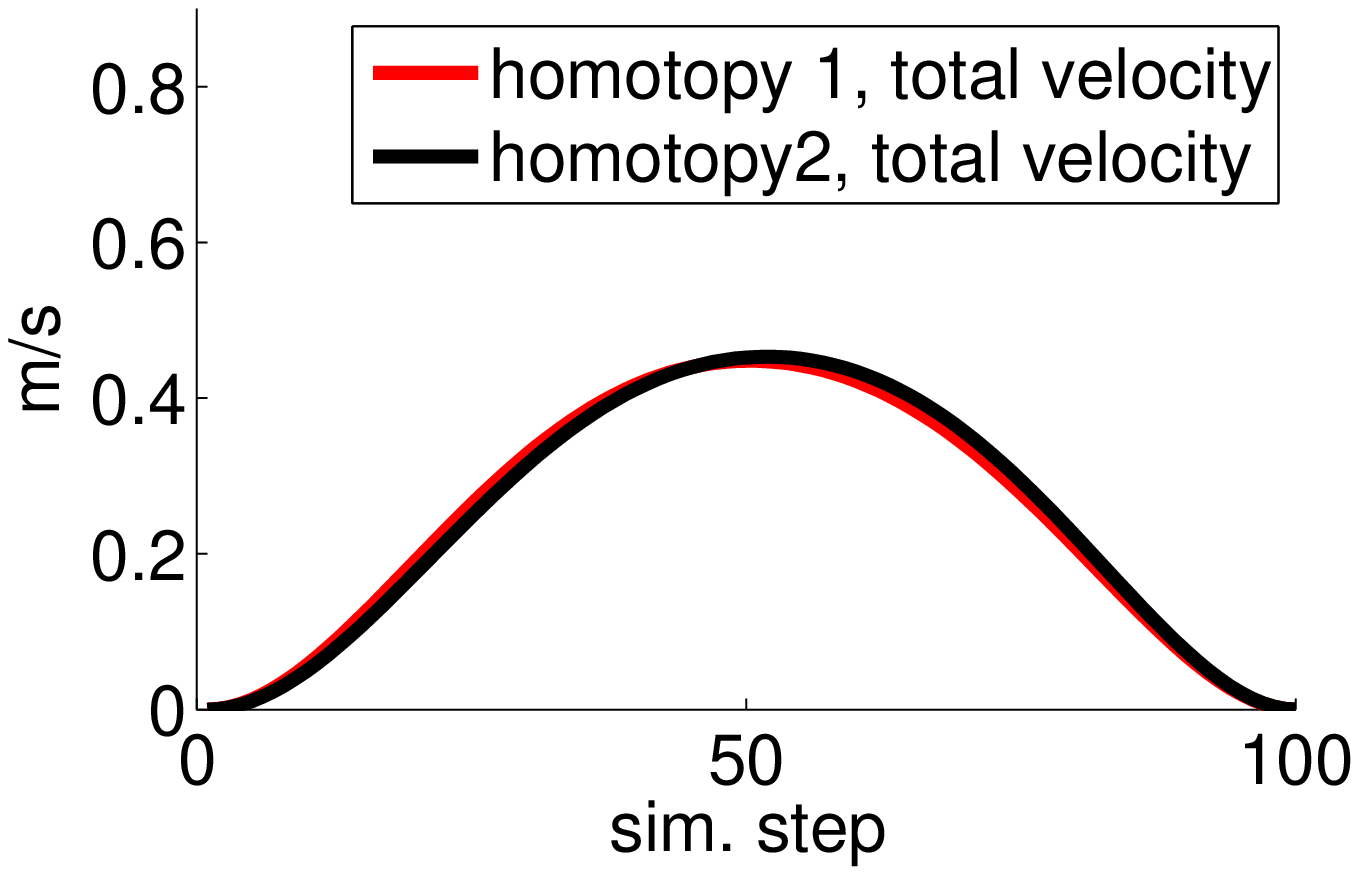}
    \label{plot1comp_vel_config1}
   }\vspace{-0.7cm}
   \subfigure[]{
    \includegraphics[width= 4.6cm, height=3.5cm] {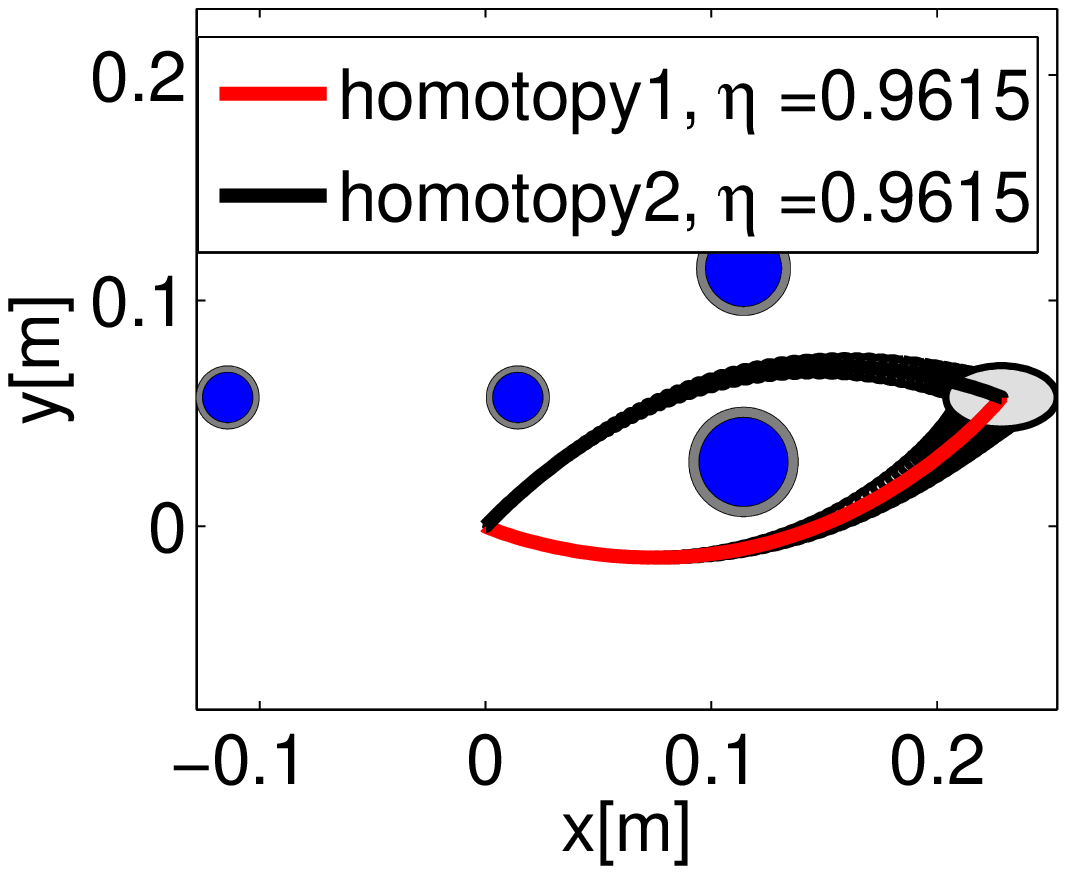}
    \label{plot2comp_config1}
   }\hspace{-0.8cm}
   \subfigure[]{
    \includegraphics[width= 4.1cm, height=3.5cm] {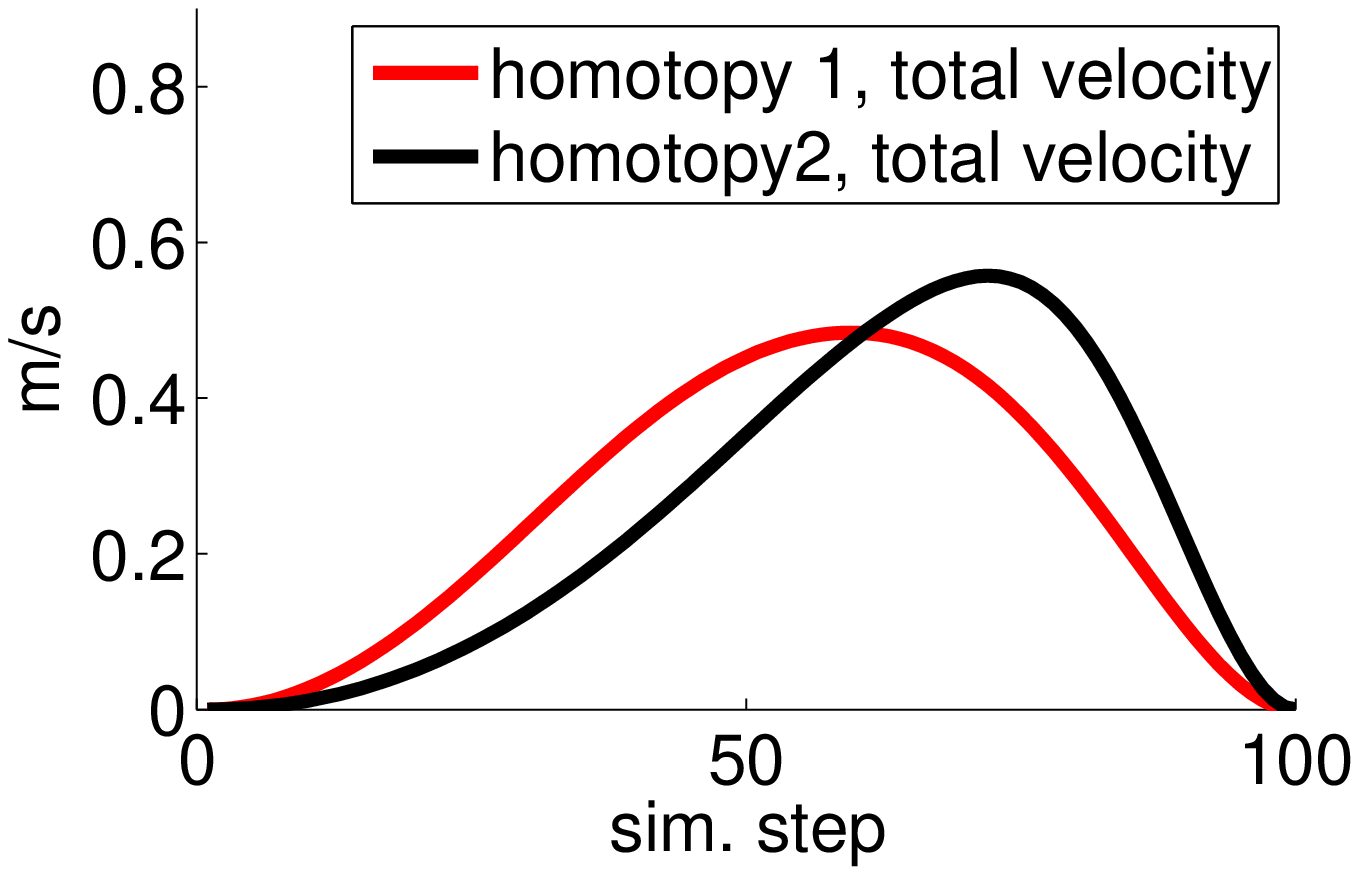}
    \label{plot2comp_vel_config1}
   }\vspace{-0.7cm}
   \subfigure[]{
    \includegraphics[width= 8.3cm, height=3.7cm] {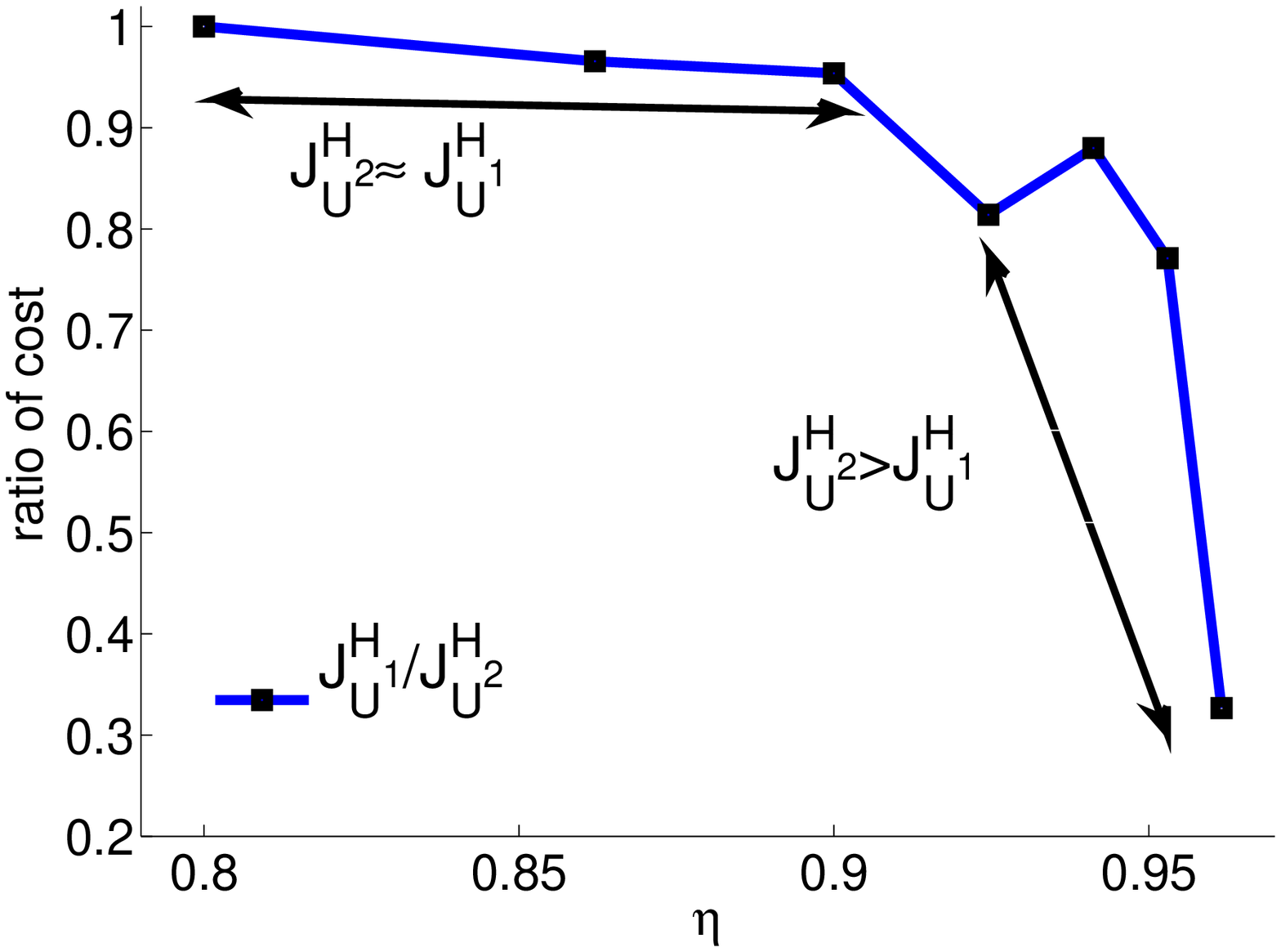}
    \label{cost_inter_hom}
   }
   \caption{Movements between the same start and goal locations and obstacle configurations but with different probability of avoidance. (a), (c) present the paths with standard deviation ellipses of the two homotopies. The obstacles are represented as blue filled circles and grey shaded region around them represent uncertainty about the size of the obstacle. (b), (d) present the velocity profiles.  (e) the ratio of the control costs between the two homotopies, $\frac{J_U^{H_1}}{J_U^{H_2}}$, as a function of $\eta$. For the chosen avoidance strategy \textbf{LV}, the control cost along the homotopies is similar for low $\eta$. For higher $\eta$, the control cost along homotopy 1 is significantly less.}        
\end{figure}

In Fig.~\ref{plot1comp_config1} and \ref{plot2comp_config1} solution trajectories of (\ref{opt2}) having same start and goal positions, but belonging to different homotopies and having different probability of avoidance, $\eta$, are depicted. The trajectories in both the homotopies were generated by choosing such values for $\tau$ and $\lambda$ that ensure collision avoidance by slowing down near the obstacles and  reducing positional variance (strategy \textbf{LV}) rather than taking a large deviation from them. Thus, as $\eta$ increases from 0.9 (figure \ref{plot1comp_config1}) to 0.965 (figure \ref{plot2comp_config1}), we observe only a small change in arc length, but a significant change in the positional variance along the trajectories.  Moreover, since trajectories of homotopy 2 move through a more cluttered environment, the reduction of positional variance along it is higher than that along trajectories of homotopy 1. 

It is possible to relate the change in positional variance as $\eta$ increases to the change in the control costs through the velocity profiles. Firstly, in contrast to Fig.~ \ref{plot1comp_vel_config1}, velocity profiles shown in Fig.~ \ref{plot2comp_vel_config1} are skewed; i.e, they have low magnitudes during the initial phases and a peak which is shifted towards the right. This is to ensure that velocity magnitudes (and thus positional variance) are low near the obstacles and reach peak only after crossing the obstacles. Since trajectories in homotopy 2 require a larger reduction in positional variance, the skewness observed in their velocity profile is also higher. Finally, the skewness in velocity profiles is accompanied with higher peak velocities. This is because of the fixed final time paradigm of the optimization, (\ref{opt2}). Since, magnitudes are low during initial phases of the trajectories, it needs to be compensated by moving faster in obstacle free space to ensure that the goal position is reached in specified time. Now, it is easy to deduce that a skewed velocity profile with higher peaks would mean higher accelerations and jerks and thus, consequently higher control costs. 

To summarize, for collision avoidance strategy \textbf{LV}, maintaining high $\eta$ requires larger reduction in positional variance leading to larger skewness in velocity profiles and consequently higher control costs. However, since trajectories in homotopy 2 require a larger reduction in positional variance, the control costs along it would increase at a higher rate than that along trajectories in homotopy 1. We demonstrate this last observation in Fig.~ \ref{cost_inter_hom} which shows the ratio of control costs along homotopy 1 and homotopy 2 for the various values of $\eta$. For lower values (till $\eta =0.9$), the cost along homotopy 1 and homotopy 2 are similar owing to their similar velocity profiles.  However, for higher $\eta$, cost along homotopy 1 is significantly lower than that along homotopy 2.

\begin{figure}[!h]
  \centering
   \subfigure[]{
    \includegraphics[width= 4.6cm, height=3.5cm] {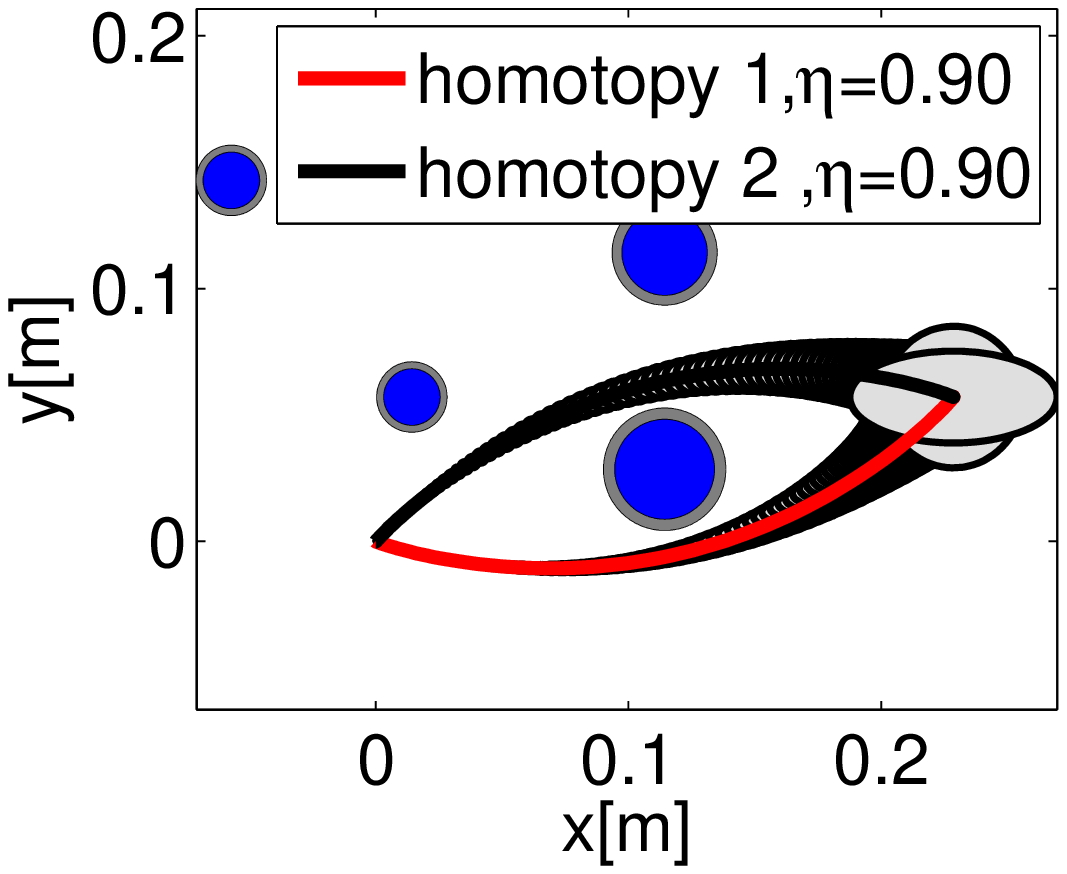}
    \label{hom_comp_traj1}
   }\hspace{-0.8cm}
   \subfigure[]{
    \includegraphics[width= 4.1cm, height=3.5cm] {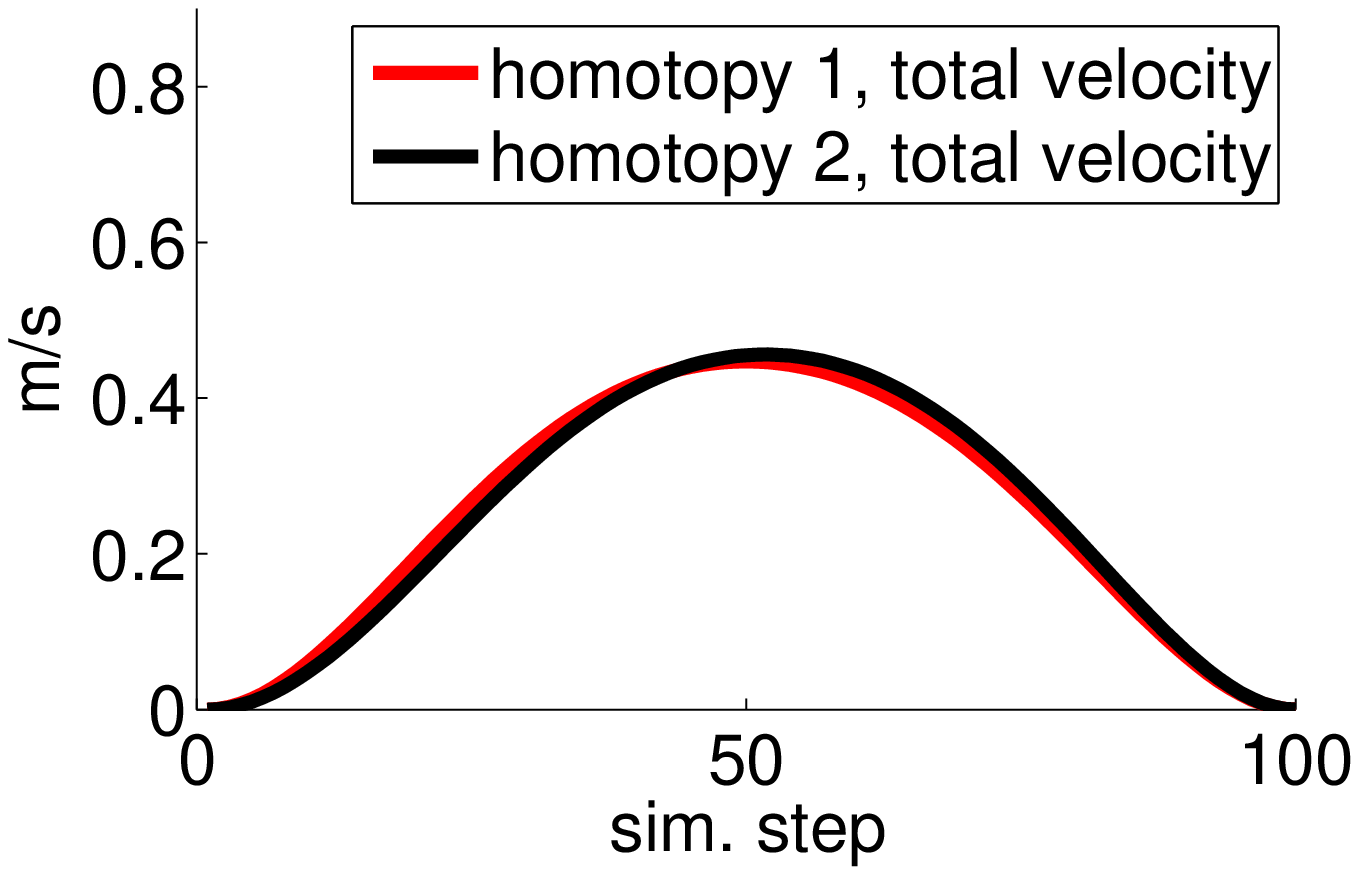}
    \label{hom_comp_traj1_vel}
   }\vspace{-0.7cm}
   \subfigure[]{
    \includegraphics[width= 4.6cm, height=3.5cm] {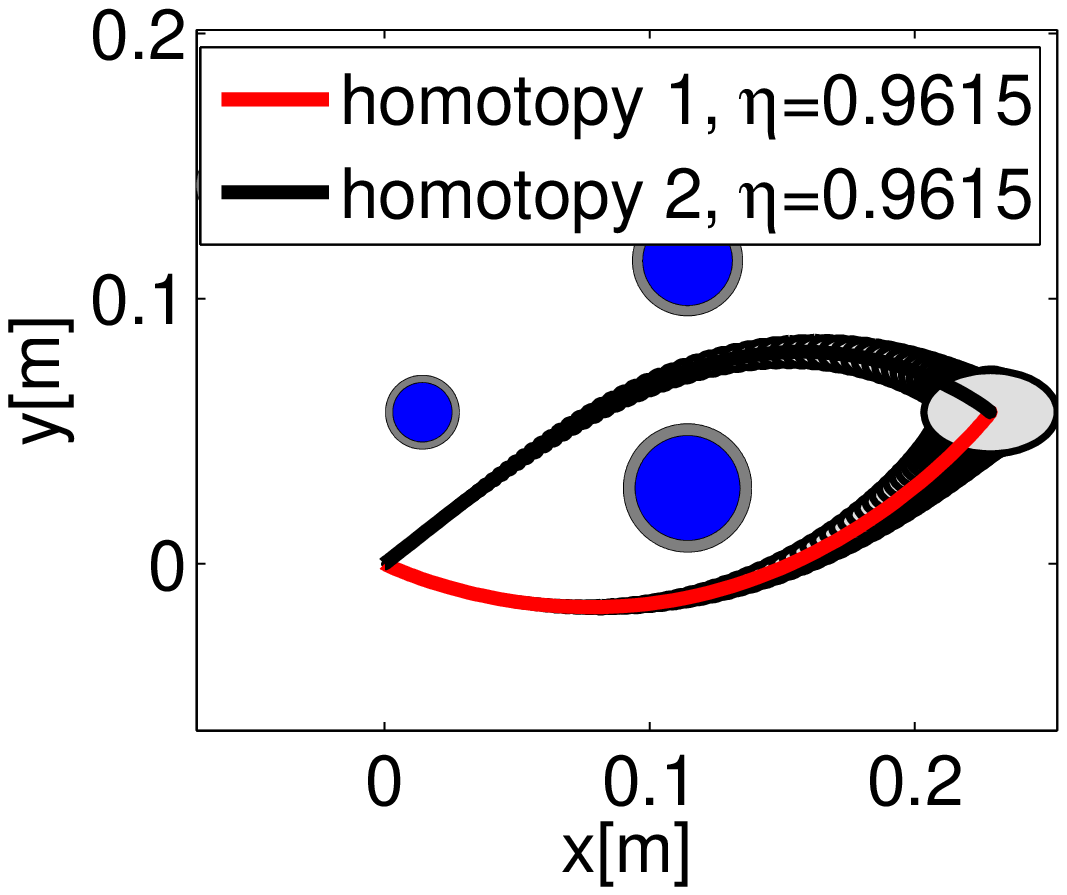}
    \label{hom_comp_traj2}
   }\hspace{-0.8cm}
   \subfigure[]{
    \includegraphics[width= 4.1cm, height=3.5cm] {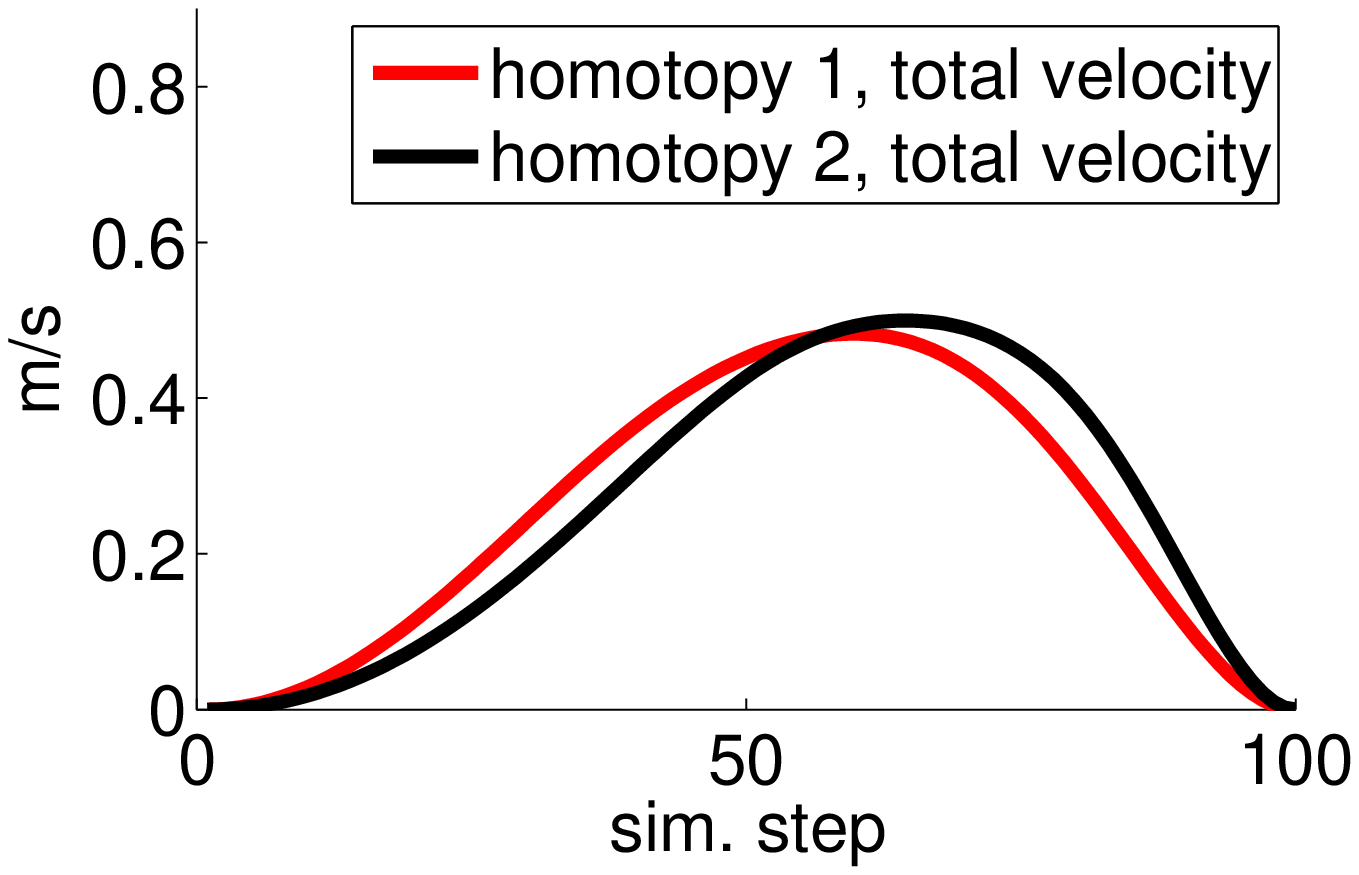}
    \label{hom_comp_traj2_vel}
   }\vspace{-0.7cm}
   \subfigure[]{
    \includegraphics[width= 8.3cm, height=3.7cm] {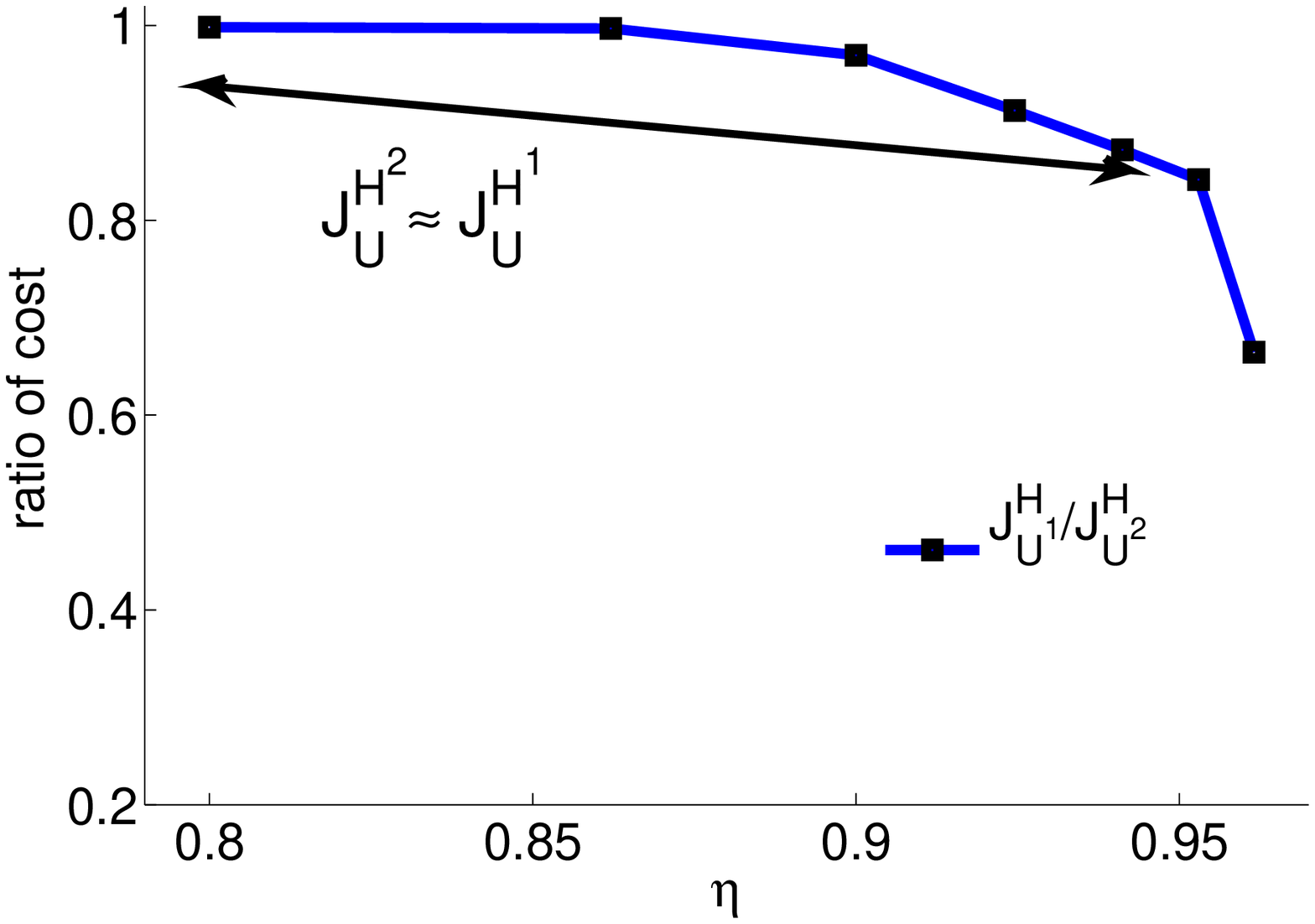}
    \label{cost_inter_hom_strat1}
   }
   \caption{Movements between the same start and goal locations and obstacle configurations but with different probability of avoidance, $\eta$ . The results are similar to that shown in Fig. 4,  but trajectories are now computed with respect to strategy \textbf{HC}, which gives higher emphasis on clearance from obstacles for obstacle avoidance. (e): Ratio of control cost along the two homotopies,$\frac{J_U^{H_1}}{J_U^{H_2}}$ with respect to strategy \textbf{HC}. }              
\end{figure}

\subsubsection{Strategy \textbf{HC}}
Here we re-analyze the cost along homotopies for the same configuration as shown in Fig.~4, but with respect to strategy \textbf{HC} where there is a bigger reliance on clearance from the obstacles for collision avoidance. 

The trajectories along both the homotopies are shown in Fig.~ \ref{hom_comp_traj1} and \ref{hom_comp_traj2}. Comparing these trajectories with Fig.~\ref{plot1comp_config1} and \ref{plot2comp_config1}, demonstrates that there is a significant increase in clearance from the obstacles with increase in $\eta$. Thus, a lesser restriction is required on the growth of positional variance and consequently, the velocity profiles along trajectories in both the homotopy become very similar even at higher $\eta$ (figure \ref{hom_comp_traj2_vel}). This is very different from the comparisons shown in Fig.~ \ref{plot2comp_vel_config1}. 

The similarity in velocity trajectories in turn results in similar control costs along both the homotopies (figure \ref{cost_inter_hom_strat1}). In particular, the control cost along homotopies 1 and 2 are similar for a larger range of $\eta$. The lowest ratio of cost is $0.67$ in figure \ref{cost_inter_hom_strat1} for $\eta = 0.9615$. In comparison, the ratio was $0.33$ in figure \ref{cost_inter_hom} for the same $\eta$.


\section{Discussion and Future Work} \label{disc}
In this paper, we presented a stochastic optimal control problem with signal dependent noise and probabilistic collision avoidance constraints as a model of human reaching among obstacles. We then reformulated it into a parameter optimization problem. The parameters $\tau$ and $\lambda$ which appeared in the reformulated optimization problem, (\ref{opt2}) served as a mapping between the probability of collision avoidance, $\eta$, and possible collision avoidance strategies. 

The parameter $\tau$ models the clearance from the obstacles, and the  parameter $\lambda$ models the effect of slowing down near the obstacles. In our simulations, we demonstrate that effect of these two parameters on movement paths and velocity profiles is in agreement with  the experimental findings reported in \cite{Tressilian} for reach to grasp movements around obstacles. Specifically, they discussed two basic but coupled strategies of collision avoidance which consists of moving around the obstacle and slowing down.

We showed how each avoidance strategy results in a unique variation of control costs  with respect to $\eta$ both within and across homotopies. These variation in control costs can be used as a basis for predicting user behavior between a given start and goal position and for a given  obstacle environment. For example, in Fig.~\ref{strat_comp_traj1}-\ref{strat_comp_traj2}, \ref{cost_intra_hom}, we showed that a risk-seeking behavior (low $\eta$) is more likely to use strategy \textbf{LV} for collision avoidance as it requires less control effort. In contrast, a risk-averse behavior (high $\eta$) would likely choose strategy \textbf{HC}.

We also showed how control cost along different homotopies is dictated by the choice of avoidance strategy. This variation in control cost can be used to predict the homotopy selection by the human. In particular, if two competing homotopies have similar control costs, then the human would have equal affinity towards either of it, thus leading to a random behavior. However, as the ratio of control costs departs from unity, the possibility of selection would  incline towards the lesser control costs, thus leading to more well defined pattern.

Our proposed framework has the following limitations. Firstly, we used a very simple dynamic model, and thus, we necessarily do not capture every aspect of the motion of the human arm. A second order linear mechanical system or a non-linear model of a serial link robotic manipulator are a better alternative. The second order mechanical system can be easily incorporated because as long as the system is linear, the structure of the optimization (\ref{opt2}) would not change. In contrast, incorporating even a simple planar two-link manipulator model is more challenging, and may require methods similar to that proposed in \cite{todorov_obstacle}. Secondly, the fixed final time paradigm of optimization (\ref{opt2}) is not equipped to capture the effect of increase in traversal time of reaching motions due to presence of obstacles. A possible solution to this could be developed using the time scaling concepts \cite{bharath_iros15}. Our current study is limited to developing and approximating an efficient solution for the computational framework, and demonstrating the homotopies and strategies that can be explained within this framework, and we do not test our predictions against real reaching data.

From our simulation study, we conclude that if the parameters $\tau$ and $\lambda$ are known, the possible choice of homotopy as well as choice of trajectory within that homotopy can be predicted. Thus, currently our efforts are focused towards developing an inverse optimization  framework which can automatically recover these parameters from example trajectories demonstrated by the human. 

\bibliographystyle{IEEEtran}
\bibliography{ref_biorobo}

\end{document}